%% file: template.tex
\documentclass{article}

\usepackage{arxiv}

\usepackage[utf8]{inputenc} 
\usepackage[T1]{fontenc}    
\usepackage{hyperref}       
\usepackage{url}            
\usepackage{booktabs}       
\usepackage{amsfonts}       
\usepackage{nicefrac}       
\usepackage{microtype}      
\usepackage{cleveref}       
\usepackage{graphicx}
\usepackage{doi}
\usepackage{tabularx}
\usepackage{authblk}
\usepackage{todonotes}
 
\title{MedAlpaca - An Open-Source Collection of Medical Conversational AI Models and Training Data}

\author[1,+]{Tianyu Han}
\author[2,+]{Lisa C. Adams}
\author[4]{Jens-Michalis Papaioannou}
\author[4]{Paul Grundmann}
\author[4]{Tom Oberhauser}
\author[4]{Alexei Figueroa}
\author[4]{Alexander Löser}
\author[1,+]{Daniel Truhn}
\author[2,5,+]{Keno K. Bressem}

\affil[1]{Department of Radiology, University Hospital Aachen, Aachen, Germany\authorcr Email: \tt \{tianyu.han, dtruhn\}@ukaachen.de \vspace{0.3cm}}
\affil[2]{Department of Diagnostic and Interventional Radiology, Technical University of Munich, School of Medicine and Health, Klinikum rechts der Isar, TUM University Hospital, Munich, Germany \authorcr Email: \tt lisa.adams@tum.de \vspace{0.3cm}}
\affil[4]{Berliner Hochschule für Technik (BHT), Berlin, Germany \authorcr Email:  \tt \{michalis.papaioannou, pgrundmann, tom.oberhauser, alexei.figueroa, aloeser\}@bht-berlin.de \vspace{0.3cm}}
\affil[5]{Department of Cardiovascular Radiology and Nuclear Medicine, Technical University of Munich, School of Medicine and Health, German Heart Center, TUM University Hospital, Munich, Germany \authorcr Email: \tt keno.bressem@tum.de \vspace{0.3cm} \vspace{0.3cm}}
\affil[+]{Contributed equally}

\begin{document}
\maketitle

\begin{abstract}
	As large language models (LLMs) like OpenAI's GPT series continue to make strides, we witness the emergence of artificial intelligence applications in an ever-expanding range of fields. In medicine, these LLMs hold considerable promise for improving medical workflows, diagnostics, patient care, and education. Yet, there is an urgent need for open-source models that can be deployed on-premises to safeguard patient privacy. In our work, we present an innovative dataset consisting of over 160,000 entries, specifically crafted to fine-tune LLMs for effective medical applications. We investigate the impact of fine-tuning these datasets on publicly accessible pre-trained LLMs, and subsequently, we juxtapose the performance of pre-trained-only models against the fine-tuned models concerning the examinations that future medical doctors must pass to achieve certification.  
 
\end{abstract}


\section{Introduction}
The advent of large language models (LLMs), trained using reinforcement learning through human feedback (RLHF) and exemplified by OpenAI's GPT series, has profoundly influenced the fields of natural language processing (NLP) and artificial intelligence (AI) research \cite{ouyang2022training}. Their remarkable capacity to produce coherent, contextually apt, and intricate responses has increased their value across diverse domains. Notably, the medical field is poised to reap substantial benefits from the implementation of these models.

A salient benefit of these LLMs lies in their ability to perform tasks following instructions in natural language, thereby eliminating the necessity for users to have programming proficiency. This feature empowers medical professionals to seamlessly engage with and steer the models through diverse medical workflows.

Potential applications include  aiding medical professionals in note-taking, composing discharge letters, retrieving information from extensive documents, summarizing content, and converting free-form texts into structured formats \cite{radiologyKeno, Sallam2023}. Provided the model has been trained on a sufficient number of medical documents, it may possess the medical knowledge necessary to assist in consultations by supplying accurate information derived from its base texts \cite{lee2023benefits}. Furthermore, the training of medical students can also benefit from these models, wherein they assume the role of a study partner, capable of quizzing students or elucidating complex subjects, provided the model demonstrates sufficient coherence and accuracy. However, the most adept LLM models are currently not openly accessible, being available exclusively through APIs that necessitate data transmission to the parent company for processing.

Considering the sensitive nature of medical data and the imperative for robust privacy safeguards, non-transparent models with unclear data management practices are ill-suited for medical applications. To tackle this challenge and avert unauthorized data transfers, it is essential to employ open-source models that enable on-site implementation, thus mitigating privacy concerns.

Addressing this demand, we present a compilation of language models specifically fine-tuned for biomedical tasks. Utilizing a blend of new and established open-source biomedical datasets, we adapt them into an instruction-following format. This structure facilitates supervised fine-tuning as the initial phase, as detailed in \cite{ouyang2022training}.

To assess the effectiveness of these models, we evaluate their performance on the United States Medical Licensing Examination (USMLE), a standardized assessment undertaken by medical students in the United States as part of their qualification process to become physicians. This evaluation offers valuable insights into the models' competencies and prospective applications within the medical domain.

We make all models and datasets publicly available, anticipating that they will confer significant advantages to both medical and AI researchers as well as practitioners in their respective fields.

\section{Materials and Methods}
\label{sec:methods}

\subsection{Datasets}
\label{sec:datasets}

In this section, we present \textbf{Medical Meadow} a collection of medical tasks that we have compiled for fine-tuning and evaluating the performance of large language models in the context of medicine. Medical Meadow consists of two main categories, a collection of established medical NLP tasks reformatted in instruction tuning formats as well as a crawl of various internet resources. Each dataset focuses on different aspects of medical knowledge and practice, providing a comprehensive training and evaluation framework.
See Table \ref{tab:models_datasets} for a detailed overview of the datasets.

\begin{table}[htbp]
  \centering
  \caption{Summary of medical datasets created for this work. For information regarding other, already published data, please refer to the respective original publication.}
  \begin{tabularx}{\linewidth}{
          p{1.75cm}
          >{\hsize=0.2\hsize}X
          >{\hsize=0.7\hsize}X
          >{\hsize=0.1\hsize}X
        }
    \toprule
    \textbf{Dataset}          & \textbf{Source}   & \textbf{Description} & \textbf{n}  \\
    \midrule
    \multicolumn{3}{c}{\textbf{Finetuning}} \\
    \midrule
    Medical Flash Cards  & \href{https://apps.ankiweb.net}{Anki} Flashcards & Rephrased Q\&A pairs derived from the front and back sides of medical flashcards & 33,955 \\
    Stack  Exchange & \href{https://data.stackexchange.com/}{Academia}  & Q\&A pairs generated from questions and their top-rated answers & 39,633\\
      & Biology                 &  & 7,482\\
      & Fitness                 &  & 3,026\\
      & Health                  &  & 1,428\\
      & Bioinformatics          &  & 906\\
    Wikidoc & \href{https://www.wikidoc.org/index.php/Main_Page}{Living Textboo}k & Q\&A pairs generated from paragraphs, where questions were formulated from rephrased paragraph titles, and answers were extracted from paragraph text & 67,704\\
      & Patient Information  & Q\&A pairs generated from paragraph headings and associated text content  & 5,942\\
    \midrule
    \multicolumn{3}{c}{\textbf{Evaluation}} \\
    \midrule    
    USMLE  & \href{https://www.usmle.org/sites/default/files/2021-10/Step_1_Sample_Items.pdf}{Step 1}  & Multiple choice questions from the USMLE self-assessment with image-based questions excluded  & 119 \\
       & \href{https://www.usmle.org/sites/default/files/2021-10/Step2_CK_Sample_Questions.pdf}{Step 2}  &  & 120 \\
       & \href{https://www.usmle.org/sites/default/files/2021-10/Step3_Sample_Items.pdf}{Step 3}  &   & 135 \\

    \bottomrule
  \end{tabularx}
  \label{tab:models_datasets}
\end{table}

\subsubsection{Dataset 1: Flash Cards Used by Medical Students}

Medicine as a whole encompasses a wide range of subjects that medical students and graduates must master in order to practice effectively. This includes a profound understanding of basic medical sciences, clinical knowledge, and clinical skills. The \href{https://apps.ankiweb.net}{Anki Medical Curriculum} flashcards are created and updated by medical students and cover the entirety of the medical school curriculum, addressing subjects such as anatomy, physiology, pathology, pharmacology, and more. These flashcards frequently feature succinct summaries and mnemonics to aid in the learning and retention of important medical concepts.
In our investigation, we leveraged flashcards as a source to create question-answer pairs for training purposes. Upon excluding cards containing images, we harnessed OpenAI's GPT-3.5-Turbo to restructure the cards into coherent, contextually pertinent question-answer pairs. Generally, the questions and answers are concise and targeted, as the flashcards offer limited space for incorporating extensive information. See Table \ref{tab:datasets_anki} for representative Q/A pairs.

\subsubsection{Dataset 2: Stackexchange Medical Sciences}
The stackexchange dataset consists of 52,475 question-answer pairs obtained from five \href{https://stackexchange.com}{Stack Exchange} forums related to biomedical sciences and related fields:
\begin{enumerate}
    \item \textit{Academia}: This forum offers insights into research methodologies, scientific publication processes, and career paths within the scientific community. While not directly affiliated with medicine, considering the volume of medical research, it is likely that medical professionals will also consult models pertaining to this subject matter.
    \item \textit{Bioinformatics}: As an interdisciplinary field combining biology, computer science, and data analysis, the Bioinformatics forum offers valuable information on the techniques and tools used for analyzing complex biological data, which is increasingly important in modern medical research.
    \item \textit{Biology}: Biology covers topics such as genetics, physiology, and molecular biology, which are all relevant to basic medical research. By including this forum, we aim to add core concepts of life sciences to the training data.
    \item \textit{Fitness}: This forum addresses the practical aspects of maintaining and improving physical health, including exercise routines, nutrition, and injury prevention. By incorporating the Fitness forum, we introduce models to health-related information that might be directly applicable to patient care and lifestyle recommendations.
    \item \textit{Health}: The Health forum covers a broad range of topics related to personal health, disease prevention, and medical treatments which could be directly transferable to medical care. 
\end{enumerate}
To maintain a high level of answer quality, we collected data exclusively from responses that received a minimum of five up-votes within the forum discussions and paired them with their corresponding questions. See Table \ref{tab:datasets_stackexchange} for representative Q/A pairs.

\subsubsection{Dataset 3: Wikidoc}

We incorporated medical question-answer pairs extracted from \href{https://www.wikidoc.org/index.php/Main_Page}{WikiDoc}, a collaborative platform for medical professionals to share and contribute up-to-date medical knowledge. The platform has two main sub-sites, the \textit{"Living Textbook"} and \textit{"Patient Information"}. The "Living Textbook" contains chapters for various medical specialties, which we crawled. We then used \texttt{GTP-3.5-Turbo} to rephrase the paragraph heading to a question and used the paragraph as answers.  Patient Information is structured differently, in that each section subheading is already a question, making rephrasing obsolete. See Table \ref{tab:datasets_wikidoc} for representative Q/A pairs.

\subsubsection{Dataset 4: medical NLP Benchmarks}

We additionally use data from open NLP datasets and benchmarks, including: 

\begin{enumerate}
    \item The COVID-19 Open Research Dataset Challenge (CORD-19), consisting of more than one million scholarly articles \cite{CORD-19}
    \item Benchmark data from Measuring Massive Multitask Language Understanding \cite{hendrycks2021ethics, hendryckstest2021}
    \item Training data from the MedQA benchmark, question answering datasets consisting of medical exam questions \cite{jin2020disease}
    \item Training data from the Pubmed Causal Benchmark \cite{yu-etal-2019-detecting}
    \item Conversational data from medical forums as presented in \cite{yunxiang2023chatdoctor}
    \item The \href{https://open-assistant.io/}{OpenAssistant} dataset. A Crowd sourced conversational dataset, especially targeted towards training models with RLHF
    
\end{enumerate}

\subsection{Model Training}
\label{sec:pretrained}
Our models are built upon the LLaMA (Large Language Model Meta AI) foundation models. LLaMA represents a cutting-edge large language model released by Meta, demonstrating their commitment to open science. It is available in various sizes, including 7 billion, 13 billion, 33 billion, and 65 billion parameters. In this study, we fine-tuned the 7 and 13 billion parameter LLaMA variants, adhering to the approach delineated by Taori et al \cite{alpaca}. 

We trained each model for five epochs, employing a learning rate of $2e^{-5}$ for the 7b model and $1e^{-5}$ for the 13b model, using a cosine learning rate scheduler. Gradient accumulation facilitated training with an effective batch size of 256. Given that this training impacts all model parameters, the hardware requirements are substantial. Consequently, we explored alternative training procedures.

First, we implemented Low-Rank Adaptation (LoRA) for weight updates to adapt the pre-trained language models to our specific tasks. LoRA is a method that involves freezing the pre-trained model weights and incorporating trainable rank decomposition matrices into each layer of the Transformer architecture \cite{hu2021lora}. This approach substantially diminishes the number of trainable parameters and GPU memory requirements for downstream tasks, making it more efficient compared to full fine-tuning and significantly reducing training time.

To further decrease memory and compute demands, we employed 8-bit matrix multiplication for the feed-forward and attention projection layers, along with an 8-bit optimizer. When combined with LoRA, this strategy further reduces the memory needed for training \cite{dettmers2022llmint8} \cite{dettmers2022optimizers}. All models trained with LoRA underwent three epochs of training at a learning rate of 2e-5.

\subsection{Evaluation Procedure}
\label{sec:evaluation}
To evaluate the performance of the fine-tuned language models, we devised an assessment methodology centered on their zero-shot performance across the United States Medical Licensing Examination (USMLE) Step 1, Step 2, and Step 3 self-assessment datasets. We excluded all questions containing images, as our primary interest lies in the models' language capabilities, and they lack visual abilities. We instructed the models to present answers in the format "Option: Answer" (e.g., "A: Penicillin"). If a model's output did not adhere to this format, they were prompted up to five times until the response was generated in the desired format. If the model failed to provide the response in the desired format, the last response was retained.

Interestingly, most of the fine-tuned models typically produced answers in the correct format after the first prompt, while only the base LLaMA models required multiple prompts. We conducted separate evaluations for each model, measuring their accuracy on the USMLE Step 1, Step 2, and Step 3 datasets individually. This approach allowed us to gain a comprehensive understanding of the models' performance across the various stages of the medical licensing examination.

\section{Results}
\label{sec:results}

Our findings on the USMLE test set are displayed in Table \ref{tab:models_results}. Fine-tuned LLMs consistently surpassed the performance of their pre-trained-only counterparts. It is worth noting that while LoRa and 8-bit fine-tuning expedited the training process, employing these methods resulted in reduced accuracy.

\begin{table}[htbp]
  \centering
  \caption{Zero shot performance on the USMLE self assessment}
    \begin{tabular}{llll}
    \hline
        \textbf{Model} & \textbf{Step1} & \textbf{Step2} & \textbf{Step3} \\ \hline
        LLaMA 7b \cite{touvron2023llama} & 0.198 & 0.202 & 0.203 \\ 
        Alpaca 7b naive \cite{alpaca} & 0.275 & 0.266 & 0.293 \\ 
        Alpaca 7b LoRA & 0.220 & 0.138 & 0.252 \\ 
        MedAlpaca 7b & 0.297 & 0.312 & 0.398 \\ 
        MedAlpaca 7b LoRA & 0.231 & 0.202 & 0.179 \\ 
        MedAlpaca 7b LoRA 8bit & 0.231 & 0.241 & 0.211 \\ 
        ChatDoctor (7b) \cite{yunxiang2023chatdoctor} & 0.187 & 0.185 & 0.148 \\ 
        LLaMA 13b \cite{touvron2023llama} & 0.222 & 0.248 & 0.276   \\ 
        Alpaca 13b naive & 0.319 & 0.312 & 0.301 \\ 
        MedAlpaca 13b & \textbf{0.473} & \textbf{0.477} & \textbf{0.602} \\ 
        MedAlpaca 13b LoRA & 0.250 & 0.255 & 0.255 \\ 
        MedAlpaca 13b LoRA 8bit & 0.189 & 0.303 & 0.289 \\ 
    \end{tabular}
  \label{tab:models_results}
\end{table}

\section{Discussion and conclusion}
\label{sec:discussion}

In this study, we introduced a novel, high-quality collection of medical text data specifically designed for training instruction-following, medical large language models (LLMs). This dataset serves as a comprehensive resource for enhancing LLM performance in the medical domain, laying the groundwork for potential integration into medical education and practice.

Using our medical text data, we fine-tuned several open-source LLM variants, adopting parameter-efficient tuning methodologies to address limited computing resources \cite{peft}. This approach is vital, as full fine-tuning of language model parameters is often unfeasible for most academic institutions. Our study demonstrates the viability of parameter-efficient fine-tuning.

We evaluated LLM performance using the United States Medical Licensing Examination (USMLE) for Steps 1, 2, and 3, which assess medical knowledge at various complexity levels. As expected, performance improved with larger pre-trained models. Applying approximation techniques, such as 8-bit precision and LoRa, during fine-tuning yielded less optimal results. However, due to considerable computational costs, we did not conduct extensive hyperparameter optimization and fine-tuning; thus, it may be possible to achieve performance comparable to vanilla-trained models through more thorough hyperparameter optimization, which we leave for future research.

The availability of additional medical datasets will likely enhance the applicability and performance of these models, creating various potential applications such as extracting structured medical information from unstructured text, supporting medical students' education through question-answering interactions to reinforce their knowledge and clarify lecture uncertainties, or assisting patients in understanding their health and improving communication between doctors and patients who often find medical language challenging.

Nevertheless, implementing LLMs for these application scenarios presents challenges and concerns. Ensuring data privacy and compliance with ethical standards is critical when handling sensitive patient data; these concerns can be addressed by deploying models locally within secure hospital networks. Moreover, models must be thoroughly evaluated and safeguarded for potential biases and inaccuracies to prevent unintended consequences in medical decision-making.

A significant limitation is LLMs' tendency to confabulate or generate text that appears plausible but is factually incorrect \cite{brown2020language}. This issue is especially concerning in the medical domain, where disseminating incorrect information can have serious implications for patient care and safety. Guaranteeing the accuracy and reliability of generated information is therefore essential, necessitating rigorous evaluation and continuous monitoring to mitigate confabulation risks and the potential harm it may cause in medical settings.

In conclusion, our work substantially contributes to the field of LLMs in medicine by providing a novel, high-quality medical dataset for research and application purposes. Further, we successfully fine-tuned and evaluated various LLMs, demonstrating that their medical domain performance increases with pre-trained model size and high-quality data availability. This progress paves the way for further exploration and development of LLMs in medicine, with potential implications for medical education, patient care, and healthcare communication.

\section{Acknowledgements}
The authors acknowledge the Scientific Computing of the IT Division at the Charité - Universitätsmedizin Berlin for providing computational resources
that have contributed to the research results reported in this paper. URL: \url{https://www.charite.de/en/research/research_support_services/research_infrastructure/science_it/#c30646061}

\bibliographystyle{ieeetr}
\bibliography{references}  

\section{Appendix}

\input{tables/anki}

\input{tables/stackexchange}

\input{tables/wikidoc}

\end{document}

%% file: tables/anki.tex
\begin{table}[htbp]
\centering
\caption{Representative question from the medical flashcards dataset.}
\begin{tabularx}{\textwidth}{>{\hsize=0.3\hsize}X|>{\hsize=0.7\hsize}X}
\toprule
\textbf{Question} & \textbf{Answer} \\
\midrule
What is the relationship between very low Mg2+ levels, PTH levels, and Ca2+ levels? & Very low Mg2+ levels correspond to low PTH levels which in turn results in low Ca2+ levels. \\
 & \\
What are the cautionary considerations and contraindications for the use of $\beta$-blockers? & $\beta$-blockers must be used cautiously in decompensated heart failure and are contraindicated in cardiogenic shock. \\
 & \\
What is the recommended next step in management for a stable child who is suspected of having intussusception and has a "target sign" on ultrasound? & The recommended next step in management for a stable child with suspected intussusception and a "target sign" on ultrasound is to undergo an ultrasound-guided air contrast enema. This procedure involves using a small amount of air and contrast material to inflate the colon and push the telescoped intestine back into place. It is a non-invasive and effective method for both diagnosing and treating intussusception in children. In some cases, the procedure may need to be repeated or followed by surgery if the intussusception cannot be reduced. It is important to treat intussusception promptly to avoid complications such as bowel obstruction, bowel perforation, or sepsis.  \\
 & \\
What condition should be suspected in a young infant who is healthy and active and has blood-tinged stools? What is the cause of this condition and how is it managed? & The presence of blood-tinged stools in a young infant who is healthy and active should raise suspicion of protein-induced proctocolitis. This condition is caused by an intolerance to milk or soy protein, which can cause inflammation in the lining of the rectum. The management of protein-induced proctocolitis involves continuing breastfeeding and advising the mother to avoid dairy and soy products in her diet. In most cases, this will lead to resolution of the symptoms within a few days to a few weeks. If the symptoms persist, further evaluation may be necessary to rule out other causes of gastrointestinal bleeding.  \\
 & \\
Which muscle abducts the arm from 0 to 15$^\circ$ and what is the name of the nerve that innervates it? & The muscle that abducts the arm from 0 to 15$^\circ$ is the supraspinatus, which is innervated by the suprascapular nerve.  \\
 & \\
What is the most common cause of community-acquired pneumonia? & Streptococcus pneumoniae is the most common cause of community-acquired pneumonia. Community-acquired pneumonia is a type of pneumonia that is acquired outside of a hospital or healthcare setting. It can be caused by a variety of bacteria, viruses, and other microorganisms, but Streptococcus pneumoniae is the most frequent cause. Other common causes of community-acquired pneumonia include Haemophilus influenzae, Mycoplasma pneumoniae, and Legionella pneumophila, among others. The specific cause of community-acquired pneumonia can vary depending on a person's age, underlying health conditions, and other factors.  \\
 & \\
What is a risk factor for pyelonephritis related to the urinary system? & One risk factor for pyelonephritis is vesicoureteral reflux.  \\
 & \\
Which type of fibrillation is a fatal arrhythmia that requires immediate CPR and defibrillation to prevent death? & Ventricular fibrillation is the type of fibrillation that is a fatal arrhythmia without immediate CPR and defibrillation.  \\
\bottomrule
\end{tabularx}
\label{tab:datasets_anki}
\end{table}

%% file: tables/stackexchange.tex
\begin{table}[htbp]
\centering
\caption{Representative question from the Stack Exchange dataset.}
\begin{tabularx}{\textwidth}{>{\hsize=0.3\hsize}X|>{\hsize=0.7\hsize}X}
\toprule
\textbf{Question} & \textbf{Answer} \\
\midrule
Given the thalidomide birth defect disaster of the 1960s, why was it later approved by the U.S. FDA for treatment of leprosy (1996) and of multiple myeloma (2006)? Presumably, the teratogenic issues and problems with permanent nerve damage didn't go away, as the liver can convert the "good" enantiomer to the "bad." Thalidomide affects even non-pregnant women, so why approve a drug with known severe complications?
   &   Because leprosy and multiple myeloma are conditions for which other treatment options are limited.
All treatments have side effects, some more debilitating than others. The trick with pharmacology is to balance the benefit of treatment with the hazards of the side effects.
Take cancer chemotherapy. The drugs we use for cancer chemotherapy are horrible poisons. We literally use derivatives of chemical weapons to treat cancer. Why do we deliberately poison cancer patients with what amounts to a weapon of mass destruction? Because if we don't, the cancer will kill them. 
In pharmacology there's a concept of a therapeutic window. That is, there's a particular amount of compound that will cause death or other severe disability. There's another (hopefully lower) level of the compound that will effectively treat the disease. In treating, we hope to hit that middle ground: enough to treat the disease, but not enough to kill the patient.
Different drugs have different therapeutic window. The window for over-the-counter drugs like asprin is rather large: the amount needed for severe side effects is many times higher than what people take to treat headaches. That's why we can allow untrained people to self-dose. (Though look at acetaminophen/paracetamol as an over-the-counter example where the window is not as large as it probably should be.) [...] \\
& \\
What is the way they determine someone died before declaring them dead?
There are stories I've heard of over the years, people who's heart stopped, were brain dead, fell under ice rivers for half an hour and miraculously survived. So there are many instances of unexplained recoveries.
How do the medical teams or doctors determine then that this person won't all of sudden get a heart beat back? 
   &   You are correct that this happens. It is infrequent (there are not "many", as you say, compared to true deaths), but it occasionally happens that someone - even in a hospital - is thought to be dead when they are not actually dead. In one of your stories, the girl was presumed to be brain dead, not dead. So strike that one. Hypothermia is a beast all unto itself, and declaring a cold person dead is a bit tricky.*
Determining death is not simple. The International Guidelines for the Determination of Death – Phase I (May 30-31, 2012) Montreal Forum Report is 46 pages long and it still doesn't have a definitive conclusion.
For the most part (and to simplify a bit), death is determined to have occurred when someone is exceedingly unlikely (determined from experience of millions of deaths) to regain function of their heart. It can be from a very wide variety of causes, but basically it follows cardiac arrest or respiratory arrest leading to cardiac arrest.
The procedure is to observe the patient carefully. In hospital, that usually includes electronic monitors of one sort or another. Out of hospital it's by observation. When there is no evidence of cardiac electrical activity capable of generating a pulse, the patient has not been breathing for some time, oxygenation of blood has fallen to beyond critical levels, and there is no neurological activity, they are pronounced dead.[...] \\
& \\
It seems there are a ton of easy ab exercises, but very few for the lower back to keep the muscles balanced as your abs get stronger. What are simple home exercises for the lower back? 
   &   Try rolling over onto your stomach and doing a "superman". You can hold tension in the position, or do reps similar to doing "reverse situps".

ref site
Also, you might think about doing slow squats or wall sits where you focus on contracting both your abs and lower back. Tense up so as to make the line between anus and belly-button as short as possible during this exercise - maybe you can replace your ab work at the same time! \\
& \\
The accepted range for the wavelengths of light that the human eye can detect is roughly between 400nm and 700nm.  Is it a co-incidence that these wavelengths are identical to those in the  Photosynthetically Active Radiation (PAR) range (the wavelength of light used for normal photosynthesis)?
Alternatively is there something special about photons with those energy levels that is leading to stabilising selection in multiple species as diverse as humans and plants?  
   &   Good question. 
If you look at the spectral energy distribution in the accepted answer here, we see that photons with wavelengths less than ~300 nm are absorbed by species such as ozone. Much beyond 750 infrared radiation is largely absorbed by species such as water and carbon dioxide. Therefore the vast majority of solar photons reaching the surface have wavelengths that lie between these two extremes.
Therefore, I would suggest that surface organisms will have adapted to use these wavelengths of light whether it be used in photoreceptors or in photosynthesis since these are the wavelengths available; i.e., organisms have adapted to use these wavelengths of light, rather than these wavelengths being special per se (although in the specific case of photosynthesis there is a photon energy sweet spot).
For example this study suggests that some fungi might actually be able to utilize ionizing radiation in metabolism. This suggests that hypothetical organisms on a world bathed in ionizing radiation may evolve mechanisms to utilize this energy. \\
\bottomrule
\end{tabularx}
\label{tab:datasets_stackexchange}
\end{table}

%% file: tables/wikidoc.tex
\begin{table}[htbp]
\centering
\caption{Representative question from the Wikidoc Living Textbook and Patient Information.}
\begin{tabularx}{\textwidth}{>{\hsize=0.3\hsize}X|>{\hsize=0.7\hsize}X}
\toprule
\textbf{Question} & \textbf{Answer} \\
\midrule
What is the explanation for the epidemiology and demographics of anaphylaxis?   &   In 2006, the lifetime prevalence of Anaphylaxis was estimated to be 0.05\% to 2\% in the United States. However it is believed that this is an underestimation because the disease is underdiagnosed. The mortality rate of Anaphylaxis is approximately 186 to 225 deaths/year. Patients of all age groups may develop anaphylaxis, however, children and adolescents account for the majority of cases. There is no racial predilection to anaphylaxis.
Anaphylaxis affects men and women equally with more women being admitted to the hospital for the disease. There is no regional predilection to anaphylaxis. An estimated 1.24\% to 16.8\% of the United States population is considered at risk for developing anaphylaxis if they are exposed to one or more allergens. Anaphylaxis results in fewer than 1,000 deaths per year in the U.S and the most common presentation is cardiovascular collapse \\
 & \\
What do coronary angiographic projections refer to?   &   For the beginner angiographer the anatomic landmarks formed by the spine, catheter and diaphragm provide information to discern which tomographic view from which the image is obtained. In the LAO view the catheter and spine are seen on the right side of the image, while in the RAO they are found on the right. PA imaging places these landmarks in the center of the image. Cranial angulation can usually be distinguished from caudal angulation by the presence of the diaphragm. For cranial imaging, the patient should be asked to inspire to remove the diaphragmatic shadow from the image. \\
& \\
What are the symptoms of Hypothermia?   &   As people develop hypothermia, their abilities to think and move are often lost slowly. In fact, they may even be unaware that they need emergency treatment. Someone with hypothermia also is likely to have frostbite.
The symptoms include: \
Drowsiness Weakness and loss of coordination Pale and cold skin Confusion Uncontrollable shivering (although at extremely low body temperatures, shivering may stop) Slowed breathing or heart rate Lethargy, cardiac arrest, shock, and coma can set in without prompt treatment. Hypothermia can be fatal. \\
 & \\
 Who is at risk for hereditary pancreatitis?   &   Studies demonstrate that cationic trypsinogen gene mutations are associated with hereditary pancreatitis. The major mutations are known as cationic trypsinogen "R122H", "N29I". Further more, hereditary pancreatitis has also been linked to an increased lifetime risk of pancreatic cancer. \\
  & \\
 Who is at highest risk for Glomerular disease ?   &   The following may increase your risk of this condition:
Blood or lymphatic system disorders Exposure to hydrocarbon solvents History of cancer Infections such as strep infections, viruses, heart infections, or abscesses Many conditions cause or increase the risk for glomerulonephritis, including:
Amyloidosis Anti-glomerular basement membrane antibody disease Blood vessel diseases such as vasculitis or polyarteritis Focal segmental glomerulosclerosis Goodpasture syndrome Heavy use of pain relievers, especially NSAIDs Henoch-Schonlein purpura IgA nephropathy Lupus nephritis Membranoproliferative GN \\
& \\ 
What to expect if I have Intracranial aneurysms  (Outlook/Prognosis)?   &   The outcome varies. Patients who are in deep comas after an aneurysm rupture generally do not do as well as those with less severe symptoms. Ruptured cerebral aneurysms are often deadly. About 25\% of people die within 1 day, and another 25\% die within about 3 months. Of those who survive, about 25\% will have some sort of permanent disability. \\

\bottomrule
\end{tabularx}
\label{tab:datasets_wikidoc}
\end{table}

%% file: template.bbl
\begin{thebibliography}{10}

\bibitem{ouyang2022training}
L.~Ouyang, J.~Wu, X.~Jiang, D.~Almeida, C.~Wainwright, P.~Mishkin, C.~Zhang,
  S.~Agarwal, K.~Slama, A.~Ray, {\em et~al.}, ``Training language models to
  follow instructions with human feedback,'' {\em Advances in Neural
  Information Processing Systems}, vol.~35, pp.~27730--27744, 2022.

\bibitem{radiologyKeno}
L.~C. Adams, D.~Truhn, F.~Busch, A.~Kader, S.~M. Niehues, M.~R. Makowski, and
  K.~K. Bressem, ``Leveraging gpt-4 for post hoc transformation of free-text
  radiology reports into structured reporting: A multilingual feasibility
  study,'' {\em Radiology}, p.~230725, 2023.
\newblock PMID: 37014240.

\bibitem{Sallam2023}
M.~Sallam, ``The utility of chatgpt as an example of large language models in
  healthcare education, research and practice: Systematic review on the future
  perspectives and potential limitations,'' {\em medRxiv}, 2023.

\bibitem{lee2023benefits}
P.~Lee, S.~Bubeck, and J.~Petro, ``Benefits, limits, and risks of gpt-4 as an
  ai chatbot for medicine,'' {\em New England Journal of Medicine}, vol.~388,
  no.~13, pp.~1233--1239, 2023.

\bibitem{CORD-19}
AI2, CZI, MSR, Georgetown, NIH, and T.~W. House, ``Covid-19 open research
  dataset challenge (cord-19),'' 2019.
\newblock Kaggle Challenge.

\bibitem{hendrycks2021ethics}
D.~Hendrycks, C.~Burns, S.~Basart, A.~Critch, J.~Li, D.~Song, and
  J.~Steinhardt, ``Aligning ai with shared human values,'' {\em Proceedings of
  the International Conference on Learning Representations (ICLR)}, 2021.

\bibitem{hendryckstest2021}
D.~Hendrycks, C.~Burns, S.~Basart, A.~Zou, M.~Mazeika, D.~Song, and
  J.~Steinhardt, ``Measuring massive multitask language understanding,'' {\em
  Proceedings of the International Conference on Learning Representations
  (ICLR)}, 2021.

\bibitem{jin2020disease}
D.~Jin, E.~Pan, N.~Oufattole, W.-H. Weng, H.~Fang, and P.~Szolovits, ``What
  disease does this patient have? a large-scale open domain question answering
  dataset from medical exams,'' {\em arXiv preprint arXiv:2009.13081}, 2020.

\bibitem{yu-etal-2019-detecting}
B.~Yu, Y.~Li, and J.~Wang, ``Detecting causal language use in science
  findings,'' in {\em Proceedings of the 2019 Conference on Empirical Methods
  in Natural Language Processing and the 9th International Joint Conference on
  Natural Language Processing (EMNLP-IJCNLP)}, (Hong Kong, China), Association
  for Computational Linguistics, Nov. 2019.

\bibitem{yunxiang2023chatdoctor}
L.~Yunxiang, L.~Zihan, Z.~Kai, D.~Ruilong, and Z.~You, ``Chatdoctor: A medical
  chat model fine-tuned on llama model using medical domain knowledge,'' {\em
  arXiv preprint arXiv:2303.14070}, 2023.

\bibitem{alpaca}
R.~Taori, I.~Gulrajani, T.~Zhang, Y.~Dubois, X.~Li, C.~Guestrin, P.~Liang, and
  T.~B. Hashimoto, ``Stanford alpaca: An instruction-following llama model.''
  \url{https://github.com/tatsu-lab/stanford_alpaca}, 2023.

\bibitem{hu2021lora}
E.~J. Hu, Y.~Shen, P.~Wallis, Z.~Allen-Zhu, Y.~Li, S.~Wang, L.~Wang, and
  W.~Chen, ``Lora: Low-rank adaptation of large language models,'' {\em arXiv
  preprint arXiv:2106.09685}, 2021.

\bibitem{dettmers2022llmint8}
T.~Dettmers, M.~Lewis, Y.~Belkada, and L.~Zettlemoyer, ``Llm.int8(): 8-bit
  matrix multiplication for transformers at scale,'' {\em arXiv preprint
  arXiv:2208.07339}, 2022.

\bibitem{dettmers2022optimizers}
T.~Dettmers, M.~Lewis, S.~Shleifer, and L.~Zettlemoyer, ``8-bit optimizers via
  block-wise quantization,'' {\em 9th International Conference on Learning
  Representations, ICLR}, 2022.

\bibitem{touvron2023llama}
H.~Touvron, T.~Lavril, G.~Izacard, X.~Martinet, M.-A. Lachaux, T.~Lacroix,
  B.~Rozi{\`e}re, N.~Goyal, E.~Hambro, F.~Azhar, {\em et~al.}, ``Llama: Open
  and efficient foundation language models,'' {\em arXiv preprint
  arXiv:2302.13971}, 2023.

\bibitem{peft}
S.~Mangrulkar, S.~Gugger, L.~Debut, Y.~Belkada, and S.~Paul, ``Peft:
  State-of-the-art parameter-efficient fine-tuning methods.''
  \url{https://github.com/huggingface/peft}, 2022.

\bibitem{brown2020language}
T.~Brown, B.~Mann, N.~Ryder, M.~Subbiah, J.~D. Kaplan, P.~Dhariwal,
  A.~Neelakantan, P.~Shyam, G.~Sastry, A.~Askell, {\em et~al.}, ``Language
  models are few-shot learners,'' {\em Advances in neural information
  processing systems}, vol.~33, pp.~1877--1901, 2020.

\end{thebibliography}
